\documentclass[runningheads]{llncs}
\usepackage{graphicx}
\usepackage{amsmath,amssymb} 
\usepackage{color}
\usepackage[width=122mm,left=12mm,paperwidth=146mm,height=193mm,top=12mm,paperheight=217mm]{geometry}

\usepackage{cite}
\usepackage{amsmath,amssymb,amsfonts}
\usepackage{algorithmic}
\usepackage{graphicx}
\usepackage{textcomp}

\usepackage[inline]{enumitem}
\usepackage{algorithmic}
\usepackage{graphicx}

\usepackage{url,hyperref,microtype}
\hypersetup{
    colorlinks=true,
    citecolor=blue,
    linkcolor=blue,
    filecolor=magenta,      
    urlcolor=black
}

\usepackage{xcolor}

\usepackage{witharrows}

\usepackage[flushleft]{threeparttable}
\usepackage{tablefootnote}
\usepackage{array}

\newcolumntype{P}[1]{>{\centering\arraybackslash}m{#1}}
\usepackage{multicol}
\usepackage{multirow}
\usepackage{tabulary}
\usepackage{colortbl}
\definecolor{mygray}{gray}{0.90}
\usepackage{makecell}
\usepackage{booktabs}
\usepackage{subcaption}
\usepackage{stmaryrd}
\usepackage{float}
\usepackage{pifont}

\usepackage{arydshln}
\colorlet{mygray}{gray!15!white}

\begin{document}
\pagestyle{headings}
\mainmatter


\title{UNWIND: Any-Length Facial Video for Stress Detection without Temporal Windowing}
\titlerunning{UNWIND: Any-Length Facial Video for Stress Detection}
\authorrunning{S. Gkikas et al.}
\author{Stefanos Gkikas\inst{1} \and
Christian Arzate Cruz\inst{1} \and
Eric Nichols\inst{1} \and
Giorgos Giannakakis\inst{2} \and
Randy Gomez\inst{1}}
\institute{Honda Research Institute Japan, Wako City, Japan\\
\email{\{stefanos.gkikas, christian.arzate, e.nichols, r.gomez\}@jp.honda-ri.com}
\and
Department of Electronic Engineering, Hellenic Mediterranean University, Chania, Greece\\
\email{ggian@hmu.gr}}

\maketitle

\begin{abstract}
Automatic stress recognition from facial video provides a non-contact approach for affective monitoring. However, most existing video-based methods divide complete recordings into shorter temporal segments before performing classification. Such segmentation requires additional decisions concerning segment duration, overlap, and prediction aggregation, and may restrict the model from exploiting information distributed across the entire recording. We introduce UNWIND, a facial-video framework for stress detection that analyzes a complete recording as a single model input, eliminating the need for temporal windowing or external segmentation. UNWIND reorganizes the video by folding its temporal dimension into the channel dimension of a two-dimensional spatial representation, which is subsequently processed through a unified asymmetric-attention architecture. With a temporal stride of $\tau=1$, the framework processes the entire $120$-second sequence, corresponding to $3{,}600$ frames sampled at $30$~fps, in a single input. We evaluate seven temporal-stride settings on a stress dataset comprising $58$ subjects, using a stratified subject-level protocol that covers configurations from dense frame retention to sparse temporal sampling. The highest test accuracy, $70.02\%$, is obtained at $\tau=15$, while processing all frames at $\tau=1$ achieves a comparable accuracy of $69.73\%$. Computational requirements range from $12.48$ to $348.78$ GFLOPs across the evaluated stride settings, illustrating the balance between temporal sampling density and computational efficiency. The findings show that effective facial-video stress recognition can be achieved without dividing recordings into temporal windows and that complete-recording inference can be performed within a single unified model.
\end{abstract}

\section{Introduction}
Stress represents an integrated physiological and psychological reaction to perceived demands, involving autonomic nervous system activity and neuroendocrine responses whose magnitude and duration may vary considerably \cite{goldstein_2023, hellhammer_wust_2009}. These responses can range from short-lived reactions to specific situations to persistent chronic stress, with each form associated with different physiological patterns and potential long-term health effects. Subjective stress is commonly assessed in clinical and research contexts through questionnaire-based measures such as the Perceived Stress Scale \cite{cohen_kamarck_1983}. However, retrospective reporting can be affected by recall bias and provides limited ability to capture short-term fluctuations in stress over time \cite{shiffman_stone_2008}. Salivary cortisol is an established neuroendocrine marker of the stress response, but sample collection is intrusive, and cortisol changes occur with a temporal delay, limiting its suitability for continuous, real-time assessment \cite{hellhammer_wust_2009}.

Stress has also become an increasingly important public health concern. An analysis of nationally representative surveys from $146$ countries reported an approximately twofold increase in perceived stress across an $18$-year period, together with growing differences among demographic and socioeconomic populations \cite{canaletti_lun_2026}. Within occupational environments, psychosocial work-related factors have been associated with a quantifiable proportion of cardiovascular disease and depression cases across European countries \cite{sultantaib_villeneuve_2022}. Chronic psychological stress has additionally been associated with impaired immune regulation, increased cardiovascular risk, and depressive disorders, further emphasizing the clinical importance of dependable stress assessment \cite{cohen_janicki_2007}.

The need for reliable stress assessment is particularly evident in situations where conventional measurement approaches are difficult to apply. Self-reported assessments in clinical, workplace, and operational environments may be delayed or incomplete and can be influenced by social desirability and experimental demand effects. Wearable sensors provide an alternative means of continuously acquiring physiological information, but their practical use remains affected by user adherence, motion-related signal artifacts, and difficulties in scaling across heterogeneous deployment conditions \cite{hosseini_gottumukkala_2026}. Studies conducted outside controlled laboratory settings have likewise highlighted signal integrity and data quality as major obstacles to robust generalization \cite{neigel_vargo_2025}. These limitations strengthen the motivation for passive, non-intrusive automated approaches that infer stress objectively from signals captured naturally in everyday settings \cite{giannakakis_grigoriadis_2022}.

Facial video acquired with conventional cameras constitutes a practical non-contact modality for automated stress recognition because it does not require wearable equipment, dedicated sensing hardware, or active participation beyond remaining within camera range. Variations in facial activity associated with stress, including changes in eye behavior, mouth motion, and head movement, provide useful cues for distinguishing stress from neutral and relaxed conditions \cite{giannakakis_pediaditis_2017}. Facial Action Units further offer a structured representation of these behavioral patterns for automated analysis \cite{giannakakis_koujan_2020}. Advances in deep learning have substantially improved facial-video stress recognition, particularly through spatiotemporal models and AU-based approaches that outperform earlier handcrafted representations \cite{kyrou_kompatsiaris_2025}. Nevertheless, most existing approaches operate on fixed-duration windows or short clips rather than directly analyzing an entire recording. Consequently, videos must first be divided into shorter segments, requiring decisions about temporal partitioning and potentially losing contextual information that extends across segment boundaries \cite{zhang_feng_2020, jeon_bae_2021, valergaki_nicodemou_2026}.

We introduce UNWIND, an automatic facial-video stress detection framework that treats each complete recording as a single input, avoiding external temporal segmentation and windowing. The entire sequence is transformed through axis folding and subsequently analyzed with a unified asymmetric-attention architecture. Its internal spatial token segmentation manages the resulting high-dimensional representation without partitioning the video along the temporal dimension. Modality-agnostic Transformer architectures have also been investigated for other heterogeneous human-state recognition problems, including multimodal pain estimation from facial video and fNIRS \cite{gkikas_tsiknakis_painvit_2024} and multimodal cognitive workload assessment \cite{gkikas_workload_acii_2026}. We evaluate UNWIND on a stress dataset containing $58$ subjects using a stratified subject-level protocol and examine multiple temporal-stride settings spanning complete frame retention and progressively sparser temporal sampling. The experiments show that the same architecture can accommodate complete recordings across these configurations without structural changes.

\section{Related Work}
\label{related_work}

Research on video-based stress recognition has increasingly focused on contact-free assessment by exploiting behavioral cues conveyed through facial dynamics. Earlier approaches relied primarily on handcrafted descriptors derived from facial regions, including eye activity, mouth behavior, head-motion characteristics, and camera-based estimates of heart rate, showing that these cues can distinguish stress from neutral and anxious conditions \cite{giannakakis_pediaditis_2017}. Subsequent deep learning approaches moved beyond manually designed representations by jointly learning facial and action-related features directly from video, improving recognition performance over feature-engineering methods on dedicated stress datasets \cite{zhang_feng_2020}. Spatiotemporal models further extended this direction by learning changes in facial appearance across both space and time, typically using short, fixed-duration video clips as their basic processing unit \cite{jeon_bae_2021}. Transformer-based video architectures have also been investigated for related human-state analysis tasks, including video-based pain estimation and multimodal pain recognition combining facial recordings with heart-rate measurements \cite{gkikas_tsiknakis_embc,gkikas_tachos_2024}. More recently, researchers have examined facial-video stress analysis in naturalistic environments, where recordings are collected with fewer artificial contextual restrictions to increase ecological validity \cite{ding_xu_2025}.

Facial Action Units (AUs) offer another established representation for stress analysis by describing facial muscle activations in a physiologically interpretable form. AU-based classification methods have demonstrated that particular activation patterns contain useful information for discriminating stressed and neutral states across individuals and stress-induction conditions \cite{giannakakis_koujan_2020}. Later approaches incorporated automatic AU estimation into deep learning pipelines, combining facial geometry with deep appearance representations extracted from video for affective-state classification \cite{giannakakis_koujan_2022}. Explainable AI techniques have further supported this line of research by identifying which facial muscle activations contribute most strongly to stress predictions, providing interpretable information alongside the resulting classifications \cite{giannakakis_roussos_2025}. Graph-based formulations have additionally represented differential AU activity as interconnected graph nodes for explainable stress recognition \cite{kassiotis_stressgat_acii_2026}. Broader reviews of deep learning for stress detection identify facial-video analysis as an increasingly prominent direction, supported by continued progress in spatiotemporal representation learning \cite{kyrou_kompatsiaris_2025}. Beyond facial behavior, representation design has also played an important role in physiological stress recognition, including approaches that combine multiple image-based representations of electrodermal activity \cite{gkikas_eda_stress_prai_2026}.

Despite differences in architecture and representation, many video-based stress recognition systems divide continuous recordings into predefined temporal segments before classification. Zhang et al. \cite{zhang_feng_2020}, for instance, separated each $2$-min recording into $15$-s samples, whereas Jeon et al. \cite{jeon_bae_2021} performed stress recognition using facial clips of only $2$~s. Similar temporal decomposition remains common in more recent studies, including Transformer-based approaches based on non-overlapping windows \cite{valergaki_nicodemou_2026} and methods that first extract frame-level facial features before aggregating them in the temporal and frequency domains \cite{ding_xu_2025}. Window-based processing is also widely adopted for physiological signals, where recordings are separated into multiple segments whose learned representations are subsequently fused \cite{gkikas_kyprakis_resp_2025}. Although temporal decomposition reduces the computational burden of processing long recordings, it introduces additional design parameters, including segment duration, overlap, and prediction-aggregation strategy. These choices often depend on the dataset and experimental context and can limit access to temporal information beyond individual segment boundaries. While recent surveys highlight the expanding use of deep learning across facial, behavioral, physiological, and multimodal approaches to stress recognition \cite{kyrou_kompatsiaris_2025}, directly processing complete facial recordings without external temporal windowing remains comparatively underexplored.

\section{Methodology}
\label{sec:methodology}

\subsection{Video preprocessing}
Each recording is processed independently at its native $608\times800$ resolution (Section~\ref{ssec:data_collection}). Face localization is performed with the BlazeFace short-range detector~\cite{bazarevsky_2019_blazeface}, applied to the first frame with a detection confidence threshold of $0.5$. The returned bounding box is expanded by a $30$~pixel margin on each side and clipped to the frame boundaries. The resulting region defines the crop resolution, which is held fixed throughout the recording, so that every frame in the sequence shares identical spatial dimensions. Detection is repeated every $150$ frames, corresponding to $5$~seconds at $30$~fps, and the crop is re-anchored to the updated box position; when a re-detection returns no face, the previous box is retained, ensuring that detection dropouts do not interrupt the sequence. Between successive re-localizations, the crop position remains constant, with the $30$~pixel margin accommodating head displacement. Every frame is cropped, resampled to the fixed crop resolution, and stored at maximum quality, preserving the complete $30$~fps sequence without temporal subsampling or windowing. The face-centered frames are resized to $224\times224$ at model input.

\subsection{Video Tokenization}
UNWIND converts each facial-video recording into a token-based representation without relying on temporal windowing or modality-specific processing, allowing sequences of different durations to be handled by the same architecture. Consider a video containing $T$ frames acquired at $30$~fps. After applying a temporal stride $\tau$, the resulting sequence contains $L = \lfloor T / \tau \rfloor$ frames, where each retained frame is represented as a $224\times224$ RGB image. For a $120$-second recording with $\tau=1$, all frames are preserved, producing $L=3{,}600$ frames without temporal segmentation. UNWIND then performs \textit{axis folding}, incorporating the temporal dimension into the channel dimension. The complete sequence is therefore expressed as an $H\times W\times 3L$ tensor, maintaining the spatial arrangement of the frames while encoding temporal information along the channel axis:

\begin{equation}
\mathbf{X} \in \mathbb{R}^{B \times H \times W \times 3L},
\end{equation}

where $B$ denotes the batch size and $H=W=224$. Previous research has shown that reorganizing facial spatiotemporal information can influence recognition performance \cite{gkikas_reface_acii_2026}. Related representation-based approaches have transformed multiple physiological-signal representations into a shared image domain \cite{gkikas_kyprakis_eda_2025} and represented heterogeneous modalities through a unified tokenization strategy \cite{gkikas_arzate_pain_icmi_2026}.

Spatial information is introduced by augmenting every location $\mathbf{p}\in[-1,1]^2$ with Fourier positional features. The encoding employs $K=6$ frequency bands and a maximum frequency $f_{\max}=10$. Because the representation contains $D=2$ spatial dimensions, the positional encoding contributes $D(2K+1)=26$ additional features. It is defined as:

\begin{equation}
\gamma(\mathbf{p}) =
\bigl[
\sin(\pi s_1 \mathbf{p}),,
\cos(\pi s_1 \mathbf{p}),,
\ldots,,
\sin(\pi s_K \mathbf{p}),,
\cos(\pi s_K \mathbf{p}),,
\mathbf{p}
\bigr],
\end{equation}
where ${s_k}*{k=1}^{K}$ spans $[1,f*{\max}/2]$. The two spatial dimensions are then flattened into a sequence containing $N=H\times W=50176$ tokens. For every spatial token, the folded video channels are concatenated with the corresponding positional features, producing:

\begin{equation}
\mathbf{T} \in \mathbb{R}^{B \times N \times C'},\quad C' = 3L + D(2K+1),
\end{equation}
where $D=2$ represents the number of spatial axes. Using $D=2$ and $K=6$, the dimensionality of each token becomes:

\begin{equation}
C' = 3L + 26.
\end{equation}

The resulting token sequence is finally separated into $S=4$ consecutive spatial groups, each containing $n_s=N/S=12544$ tokens. These groups constitute spatial segments of the folded two-dimensional representation and do not correspond to temporal windows in the original video. The token set associated with spatial segment $s$ is denoted by $\tilde{\mathbf{T}}_s \in \mathbb{R}^{B\times n_s\times C'}$.

\subsection{Asymmetric Attention}
\label{sec:asymmetric}

The tokenized representation is processed by four successive layers, each consisting of one cross-attention module followed by $R_\ell$ self-attention modules. Each spatial segment is assigned a single latent state. At runtime, these states are initialized by replicating a common vector obtained from $M_0=32$ learnable global latent parameters ${\boldsymbol{\ell}*m}*{m=1}^{M_0}$:

\begin{equation}
\boldsymbol{\ell}*{\mathrm{init}} = \frac{1}{M_0}\sum*{m=1}^{M_0}
\boldsymbol{\ell}_m \in \mathbb{R}^{d_0},
\end{equation}
where $d_0=128$. The latent states associated with individual segments are therefore not independently optimized parameters. Instead, their segment-dependent representations are progressively formed through the subsequent attention operations.

\textbf{Cross-attention.} Within layer $\ell$, the latent state assigned to each spatial segment retrieves information only from the token group corresponding to that segment:

\begin{equation}
\mathbf{e}_s^{(\ell)} = \mathbf{e}_s^{(\ell-1)} +
\mathrm{Attn}\bigl(\mathbf{e}_s^{(\ell-1)},\ \tilde{\mathbf{T}}_s\bigr),
\end{equation}

where $\mathbf{e}*s^{(\ell-1)}\in\mathbb{R}^{B\times 1\times d*\ell}$ acts as the query representation, while $\tilde{\mathbf{T}}*s\in\mathbb{R}^{B\times n_s\times C'}$ supplies the keys and values. The attention mechanism is asymmetric because the query side contains only one vector of dimension $d*\ell$, whereas the key-value side contains $n_s \gg 1$ token vectors of dimension $C'$. Consequently, the attention map has dimensions $1\times n_s$ rather than forming a square matrix, and the query and key-value representations differ both in sequence length and feature dimensionality. Computation for all $S$ segments is performed in parallel by arranging the segment dimension within the batch dimension, without modifying the attention operation itself. Cross-attention employs one head in every layer, with head dimensions ${64, 48, 32, 16}$ across the four layers.

\textbf{Self-attention.} Following cross-attention, the latent states from all spatial segments are combined into the matrix $\mathbf{E}^{(\ell)} \in \mathbb{R}^{B\times S\times d_\ell}$. Self-attention is then performed jointly over the $S$ segment states, allowing information captured by different spatial regions to interact globally:

\begin{equation}
\mathbf{E}^{(\ell)} \leftarrow \mathbf{E}^{(\ell)} +
\mathrm{Attn}\bigl(\mathbf{E}^{(\ell)},\ \mathbf{E}^{(\ell)}\bigr),
\end{equation}
This operation is repeated $R_\ell \in {8, 6, 4, 2}$ times in layers $\ell = 0,\ldots,3$. Multi-head attention is used for these self-attention operations, with ${8, 6, 4, 2}$ heads and corresponding head dimensions of ${64, 48, 32, 16}$ across successive layers.

\begin{figure*}
\begin{center}
\includegraphics[scale=0.135]{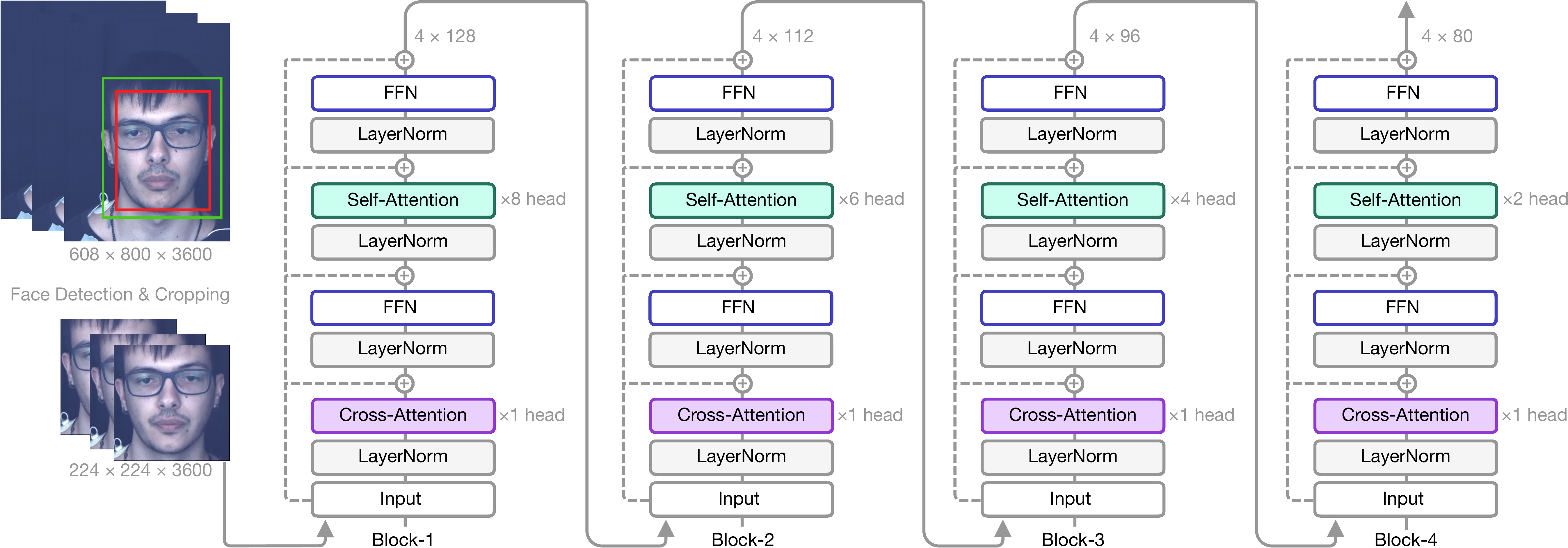}
\end{center}
\caption{Overview of UNWIND. The detected face box (red) is expanded by a $30$~pixel margin (green) and the crop is resampled to $224\times224$; at $\tau=1$ all $3{,}600$ frames of the $120$-second recording are retained. The sequence is folded along the channel axis, tokenized, and partitioned into $S=4$ spatial segments, then processed by four blocks. Each block applies cross-attention from the segment states to their corresponding token group, followed by a self-attention sub-block, repeated $R_\ell \in \{8, 6, 4, 2\}$ times, that exchanges information across all segment states; annotated head counts are per sub-block. The segment-state matrix shown above each block contracts from $4\times128$ to $4\times80$ across the four blocks.}
\label{model}
\end{figure*}

\textbf{Hierarchical segment-state compression.} The dimensionality of the segment states is progressively reduced through the four layers according to $d_\ell \in {128, 112, 96, 80}$. Whenever the dimensionality changes between adjacent layers, a linear projection maps the segment representation to the required feature size. The number of latent segment states remains unchanged at $S=4$ throughout the network, preserving one state for each spatial token group. After the fourth layer, the resulting states $\mathbf{E}^{(4)} \in \mathbb{R}^{B\times S\times 80}$ are averaged across the segment dimension and supplied to a linear classification head. Cross-attention and self-attention both employ pre-layer normalization and residual connections, while attention and feedforward dropout are fixed at $0.10$ throughout the architecture.
Table~\ref{tab:architecture} summarizes the architectural configuration, while Fig.~\ref{model} illustrates the complete processing pipeline and the organization of the four attention blocks.

\begin{table}
\caption{Architectural hyperparameters of UNWIND.}
\label{tab:architecture}
\begin{center}
\scriptsize
\begin{threeparttable}
\begin{tabular}{>{\raggedright\arraybackslash}m{5.5cm} P{2.0cm}}
\toprule
\textbf{Hyperparameter} & \textbf{Value} \\
\midrule
\midrule
Depth                                      & 4                \\\hdashline
Latent pool size ($M_0$)                   & 32               \\\hdashline
Latent dimension ($d_\ell$)                & 128, 112, 96, 80 \\\hdashline
Cross-attention heads                      & 1, 1, 1, 1       \\\hdashline
Cross-attention head dimension             & 64, 48, 32, 16   \\\hdashline
Self-attention heads                       & 8, 6, 4, 2       \\\hdashline
Self-attention head dimension              & 64, 48, 32, 16   \\\hdashline
Self-attention blocks per cross ($R_\ell$) & 8, 6, 4, 2       \\\hdashline
Spatial segments ($S$)                     & 4                \\\hdashline
Attention dropout                          & 0.10             \\\hdashline
Feedforward dropout                        & 0.10             \\\hdashline
Fourier frequency bands ($K$)              & 6                \\\hdashline
Maximum frequency ($f_{\max}$)             & 10               \\
\bottomrule
\end{tabular}
\begin{tablenotes}[para,flushleft]
\scriptsize
\item Per-layer values are listed from layer~$1$ to layer~$4$.
\end{tablenotes}
\end{threeparttable}
\end{center}
\end{table}

\section{Experimental Evaluation \& Results}

We evaluate UNWIND under several temporal-stride configurations within a binary classification framework. For the validation set, we quantify performance using macro-averaged accuracy, precision, and F1 score, whereas we assess test-set performance using macro-averaged accuracy.

\subsection{Dataset and Protocol}
\label{ssec:data_collection}
The study employed a stress dataset containing $58$ adults ($24$ men, $34$ women) with a mean age of $26.9\pm4.8$ years. The experimental procedure consisted of four stress-induction phases: social exposure, emotional recall, mental workload, and stressful video stimuli. Each participant completed $11$ tasks, including $4$ neutral, $6$ stress-inducing, and $1$ relaxation task, as reported in Table~\ref{tbl:tasks}. During the social exposure phase, a psychologist conducted an interview focused on negative personality traits. In the emotional recall phase, participants were asked to relive a previous stressful experience in real time. Mental workload was elicited using a modified Stroop Color-Word Test \cite{stroop_1935} and the Paced Auditory Serial Addition Test \cite{tombaugh_2006}. During the stressful stimuli phase, participants watched videos depicting accidents and acrophobia, while a relaxing video was used between induction phases as a physiological recovery baseline to reduce carryover effects. Heart Rate measurements confirmed the effectiveness of the stress induction, showing a statistically significant increase during stress tasks ($p<0.05$). Subjective assessment using the Self-Assessment Manikin further validated these effects, with significantly higher arousal and lower valence reported during stressful phases relative to neutral baselines.

Facial video was acquired at $60$~fps at $608\times800$ pixels and subsequently subsampled to $30$~fps. ECG was continuously recorded from a single channel at $1$~kHz. This study uses facial video as the only input modality. We used binary classification to discriminate neutral from stress conditions, with the relaxation task assigned to the neutral category. The local Research Ethics Committee approved the study (approval no.~155/12-09-2022), and informed consent was obtained from all participants. The dataset is available for non-commercial research upon request.
\footnote{\url{https://github.com/ggian/stress_dataset}}

The subjects are divided into training, validation, and testing sets at the participant level, preventing any individual from appearing in more than one subset. To reduce possible performance inflation from variations in subject difficulty, we apply a stratified splitting protocol. Leave-one-subject-out cross-validation is performed across all recorded modalities to estimate the difficulty of each subject, producing a ranking that is not determined by any single signal source. Subjects are ordered according to the combined z-score and subsequently divided into four quartiles. The resulting partition includes $38$ training, $8$ validation, and $12$ testing subjects, with each subset proportionally representing all four groups. Table~\ref{tab:subject_split_stress} provides the complete subject-level partition to support reproducibility and direct comparison with future studies.

\begin{table}
\caption{Experimental tasks employed in this study.}
\label{tbl:tasks}
\begin{center}
\scriptsize
\begin{threeparttable}
\begin{tabular}{P{0.5cm} P{3.5cm} P{2.0cm} P{1.0cm}}
\toprule
\# & Task & Duration (sec) & State \\
\midrule
\midrule
\multicolumn{4}{l}{\textit{Social Exposure}} \\
1  & Neutral reference        & 120 & N \\\hdashline
2  & Baseline description     & 120 & N \\\hdashline
3  & Interview                & 120 & S \\
\midrule
\multicolumn{4}{l}{\textit{Emotional Recall}} \\
4  & Neutral reference        & 120 & N \\\hdashline
5  & Recall stressful event   & 120 & S \\
\midrule
\multicolumn{4}{l}{\textit{Mental Workload}} \\
6  & Reading reference        & 120 & N \\\hdashline
7  & Stroop Colour-Word Test  & 120 & S \\\hdashline
8  & PASAT task               & 120 & S \\
\midrule
\multicolumn{4}{l}{\textit{Stressful Stimuli}} \\
9  & Relaxing video$^{*}$     & 120 & R \\\hdashline
10 & Adventure video          & 120 & S \\\hdashline
11 & Psychological pressure   & 120 & S \\
\bottomrule
\end{tabular}
\begin{tablenotes}[para,flushleft]
\scriptsize
\item N\,=\,neutral\quad S\,=\,stress \quad R\,=\,relaxed. *: Used as a physiological recovery baseline between induction phases; grouped with the neutral tasks for binary classification.
\end{tablenotes}
\end{threeparttable}
\end{center}
\end{table}

\begin{table}
\caption{Subject-level split by difficulty group. Subjects are ranked by
combined z-score and assigned to four quartile-based groups
(Q1\,=\,hardest, Q4\,=\,easiest).}
\label{tab:subject_split_stress}
\begin{center}
\setlength{\tabcolsep}{3pt}
\begin{threeparttable}
\scriptsize
\begin{tabular}{P{1.35cm} P{2.4cm} P{2.4cm} P{2.4cm} P{2.4cm}}
\toprule
\multirow[c]{3}{*}{Split} &
\multicolumn{4}{c}{Difficulty Group} \\
\cmidrule(lr){2-5}
 & Q1 -- Hard & Q2 -- Med-Hard & Q3 -- Med-Easy & Q4 -- Easy \\
\midrule
\midrule
Training (38) &
P017, P018, P022, P026, P034, P035, P042, P045, P050, P056 &
P001, P002, P003, P007, P012, P021, P033, P040, P048 &
P004, P014, P016, P032, P036, P046, P047, P052, P053, P054 &
P005, P010, P020, P028, P029, P037, P039, P041, P057 \\\hdashline
Validation (8) &
P038, P055 &
P009, P023 &
P006, P013 &
P019, P030 \\\hdashline
Testing (12) &
P008, P025, P044 &
P011, P024, P043 &
P015, P031, P058 &
P027, P051, P059 \\
\bottomrule
\end{tabular}
\begin{tablenotes}[para,flushleft]
\scriptsize
\item Q1: $z < -0.46$;\quad Q2: $-0.46 \leq z < -0.05$;\quad
Q3: $-0.05 \leq z < +0.40$;\quad Q4: $z \geq +0.40$.
\end{tablenotes}
\end{threeparttable}
\end{center}
\end{table}

\subsection{Video}
\label{sec:video}
\begin{figure}
\begin{center}
\includegraphics[scale=0.56]{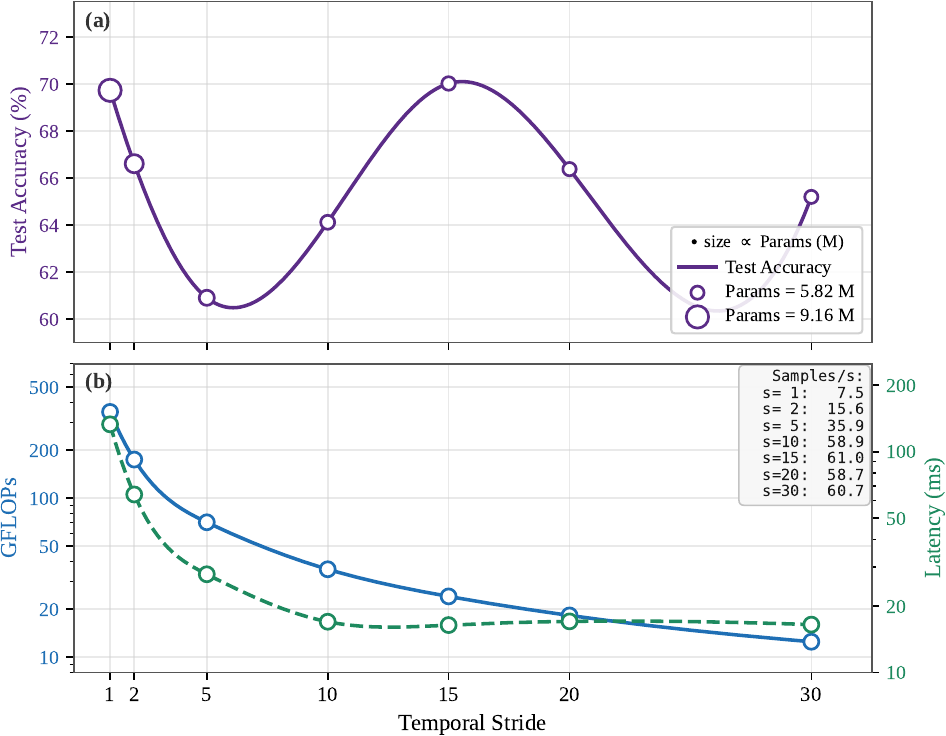}
\end{center}
\caption{Performance and computational requirements of UNWIND under different temporal-stride settings.
(a)~Test accuracy and model parameter count versus stride $\tau$; circle size represents the number of parameters, ranging from $5.82$ M at $\tau=30$ to $9.16$ M at $\tau=1$. (b)~Computational complexity in GFLOPs (solid blue, left axis) and inference latency in milliseconds (dashed green, right axis) across stride values, both displayed on logarithmic scales; the corresponding throughput in samples per second is indicated for each configuration. Inference measurements were obtained using an NVIDIA A100 GPU.}
\label{performances}
\end{figure}

Table~\ref{table:videos} summarizes the classification performance and computational requirements for all seven temporal stride settings, while Fig.~\ref{performances} jointly illustrates the corresponding accuracy and efficiency behavior. With the most temporally dense configuration ($\tau=1$), the complete $120$-second recording is preserved at $30$~fps, resulting in $L=3{,}600$ frames and a token channel dimensionality of $C'=10{,}826$. Among the evaluated settings, this configuration requires the most parameters ($9.16$M), the greatest computational workload ($348.78$ GFLOPs), and the highest inference latency ($133.02$ ms). Its resulting throughput is $7.52$ samples per second, while its test accuracy reaches $69.73\%$.

As the temporal stride increases, $L$ decreases proportionally, thereby reducing both the token channel dimension $C' = 3L + 26$ and the size of the corresponding input projection weights. At $\tau=30$, the sequence contains only $120$ retained frames, decreasing the parameter count to $5.82$ M, the computational cost to $12.48$ GFLOPs, and the inference latency to $16.47$ ms. This corresponds to an approximately $28$-fold decrease in computation compared with $\tau=1$. Fig.~\ref{performances}(b) shows that GFLOPs decline substantially with increasing stride, whereas latency decreases sharply until $\tau=10$ before exhibiting more moderate changes. Throughput follows the opposite trend, rising rapidly for the smaller stride values and subsequently stabilizing at approximately $59$--$61$ samples per second for $\tau \geq 15$.

The relationship between temporal stride and test accuracy is non-monotonic. The best test performance occurs at $\tau=15$, reaching $70.02\%$, followed by the full-frame configuration at $\tau=1$ with $69.73\%$. The highest validation accuracy is obtained at $\tau=20$ ($70.87\%$). In contrast, $\tau=5$ produces the lowest test accuracy of $60.91\%$, despite achieving the second-highest validation accuracy of $69.11\%$, indicating variability in generalization across stride settings for the $12$-subject test partition. Test accuracy for the other configurations ranges from $64.12\%$ to $66.61\%$. Overall, these findings suggest that increasing temporal sampling density beyond a moderate level does not consistently improve discriminative performance, while UNWIND supports the entire evaluated stride range without requiring any architectural modification.

\begin{table}
\caption{Performance and computational cost using the video modality.}
\label{table:videos}
\begin{center}
\setlength{\tabcolsep}{3pt}
\begin{threeparttable}
\tiny
\begin{tabular}{P{1.05cm} P{0.8cm} P{1.0cm} P{1.0cm} P{0.95cm} P{1.15cm} P{1.05cm} P{1.15cm} P{0.5cm} P{1.20cm}}
\toprule
\multicolumn{2}{c}{Input} &
\multicolumn{2}{c}{\makecell{Computational\\Cost}} &
\multicolumn{2}{c}{\makecell{Inference\\Cost}} &
\multicolumn{3}{c}{Validation} &
\multicolumn{1}{c}{Testing} \\
\cmidrule(lr){1-2}\cmidrule(lr){3-4}\cmidrule(lr){5-6}\cmidrule(lr){7-9}\cmidrule(lr){10-10}
Modality & Stride & \makecell{Params\\(M)} & GFLOPs &
\makecell{Latency\\(ms)$\downarrow$} & \makecell{Samples/s\\$\uparrow$} &
Accuracy & Precision & F1 & Accuracy \\
\midrule
\midrule
Video & 1  & 9.16 & 348.78 & 133.02 & 7.52  & 64.28 & 67.09 & 60.89 & 69.73 \\\hdashline
Video & 2  & 7.43 & 174.83 & 64.00  & 15.63 & 66.79 & 67.49 & 66.91 & 66.61 \\\hdashline
Video & 5  & 6.39 & 70.46  & 27.84  & 35.92 & 69.11 & 69.77 & 67.81 & 60.91 \\\hdashline
Video & 10 & 6.05 & 35.67  & 16.97  & 58.92 & 66.76 & 66.58 & 66.53 & 64.12 \\\hdashline
Video & 15 & 5.93 & 24.07  & 16.39  & 61.01 & 68.16 & 68.73 & 68.28 & 70.02 \\\hdashline
Video & 20 & 5.87 & 18.27  & 17.03  & 58.72 & 70.87 & 72.24 & 68.91 & 66.38 \\\hdashline
Video & 30 & 5.82 & 12.48  & 16.47  & 60.71 & 68.97 & 70.30 & 69.12 & 65.20 \\
\bottomrule
\end{tabular}
\begin{tablenotes}[para,flushleft]
\scriptsize
\item Inference Cost measured on an NVIDIA A100 GPU.
\end{tablenotes}
\end{threeparttable}
\end{center}
\end{table}


\subsection{Comparison with a Related Approach}

Table~\ref{table:comparison} compares the proposed approach at a stride of $\tau=15$ with the study in \cite{gkikas_eda_stress_prai_2026}, where electrodermal activity was employed for stress detection on the same dataset and under the same stratified hold-out protocol.
In terms of recognition accuracy, the two approaches are closely matched, reaching $70.02\%$ for the facial video and $70.97\%$ for the electrodermal activity. The computational requirements, however, differ substantially, since the proposed model comprises $5.93$ million parameters and requires $24.07$ GFLOPs, whereas the electrodermal activity model comprises $2.06$ million parameters and requires $1.58$ GFLOPs. This difference is expected, given the considerably larger volume of information in a facial video than in a single physiological channel. Nevertheless, it remains a consideration that should not be overlooked in real-world deployments, particularly in settings where inference speed and energy consumption are subject to strict constraints.

We additionally evaluated UNWIND using a leave-one-subject-out (LOSO) protocol while keeping the model architecture and training procedure unchanged. LOSO provides a more clinically oriented evaluation, as each prediction is made for an individual excluded from model training. It also facilitates comparison with future studies that adopt the same protocol. Under this protocol, UNWIND achieved a mean accuracy of $80.04\%$ (SD, $10.86\%$) across subjects. The corresponding standard deviations for precision and F1-score were $10.03\%$ and $11.52\%$, respectively.

\begin{table}
\scriptsize
\caption{Comparison of performance and computational cost with a related stress-detection approach.}
\label{table:comparison}
\begin{center}
\begin{threeparttable}
\begin{tabular}{P{1.3cm} P{1.2cm} P{1.6cm} P{1.5cm} P{2.5cm} P{1.5cm}}
\toprule
\multirow[c]{3}{*}{Study} &
\multirow[c]{3}{*}{Modality} &
\multicolumn{2}{c}{Computational Cost} &
\multirow[c]{3}{*}{Validation Protocol} &
\multirow[c]{3}{*}{Accuracy} \\
\cmidrule(lr){3-4}
& & Params (M) & GFLOPs & & \\
\midrule
\midrule

Ours & Video & 5.93 & 24.07  & Hold-out & 70.02 \\\hdashline
\cite{gkikas_eda_stress_prai_2026} & EDA & 2.06 & 1.58  & Hold-out & 70.97 \\\midrule

Ours & Video & 5.93 & 24.07 & LOSO & $80.04 \pm 10.86$ \\

\bottomrule
\end{tabular}

\begin{tablenotes}[para,flushleft]
\tiny
Hold-out refers to the same stratified hold-out protocol described in this study.
For LOSO, the standard deviations for accuracy, precision, and F1-score are 10.86, 10.03, and 11.52, respectively.
\end{tablenotes}

\end{threeparttable}
\end{center}
\end{table}

\subsection{Overall Analysis \& Discussion}

This evaluation demonstrates that UNWIND can process an entire facial recording as a single input, avoiding both temporal windowing and external segmentation. With $\tau=1$, the model receives all $3{,}600$ frames of a $120$-second recording in one forward pass. This contrasts with conventional video-based stress-recognition approaches, which typically divide recordings into short clips or fixed temporal intervals before classification. UNWIND instead represents the complete sequence through axis folding and applies asymmetric attention, while its internal spatial token partitioning remains independent of any temporal division of the video.

The temporal-stride experiments further demonstrate that this property extends beyond the highest sampling density. Without modifying its architecture, UNWIND processes sequences ranging from $120$ frames at $\tau=30$ to $3{,}600$ frames at $\tau=1$, while the token channel dimension $C' = 3L + 26$ varies according to the number of retained frames. The framework does not employ clip-level aggregation, temporal pooling, or any external segmentation mechanism. These characteristics align with UNWIND's primary objective: supporting facial videos of arbitrary length across different temporal-stride configurations without requiring structural changes to the model.

The highest test accuracy is achieved at $\tau=15$ ($70.02\%$) with a computational cost of $24.07$ GFLOPs, substantially lower than the $348.78$ GFLOPs required at $\tau=1$. Nevertheless, the full-frame setting remains competitive, obtaining $69.73\%$ test accuracy while processing the complete $3{,}600$-frame sequence. As illustrated in Fig.~\ref{performances}(a), test performance varies non-monotonically with temporal stride, showing that retaining frames at a higher temporal density does not inherently lead to better generalization. Moderate temporal subsampling can therefore maintain facial information relevant to stress recognition while considerably lowering computational requirements.

Nevertheless, the present study has a few limitations. The evaluation is restricted to a single dataset and a binary classification setting that distinguishes neutral from stress conditions. In addition, we do not directly compare UNWIND with video window-based approaches under the same validation. Therefore, although the results demonstrate the feasibility of directly processing complete recordings of several seconds and thousands of frames, they should not be interpreted as evidence that full-recording inference universally outperforms or should replace window-based strategies for video stress recognition.


\section{Conclusion}

This study introduced UNWIND, a framework for automatic stress recognition from facial video that processes complete recordings without relying on temporal windowing or external segmentation. The proposed method folds the temporal dimension into the channel dimension of a two-dimensional spatial representation and applies asymmetric attention to the resulting high-dimensional input through a compact collection of segment states. This design lets the same architecture handle inputs ranging from sparse temporal subsampling to complete frame sequences without structural changes. Experiments conducted on a stress dataset comprising $58$ subjects showed that UNWIND can process an entire $120$-second recording at $\tau=1$, corresponding to $3{,}600$ frames retained at $30$~fps, while also supporting computationally lighter stride configurations. The highest test accuracy of $70.02\%$ was achieved at $\tau=15$, whereas the full-frame configuration remained competitive with an accuracy of $69.73\%$. These results show that effective facial-video stress recognition can be achieved without dividing recordings into predefined temporal windows. Instead, the complete sequence can be processed within a single unified architecture, avoiding additional decisions related to window duration, overlap, and prediction aggregation while preserving information across the full recording. Future work should investigate longer and more variable-duration recordings, where the ability to process complete sequences may become increasingly important, and should include direct comparisons with window-based baselines using the same subject-level partition. Further extensions to additional modalities and multi-class recognition settings are also relevant directions, along with methods for identifying the portions of a recording that contribute most strongly to the final prediction, particularly in real-world monitoring scenarios.

\section*{Acknowledgments}

The authors used large language model (LLM)-based tools for language editing and improvement. All scientific content, results, and conclusions are solely the work of the authors.


\bibliographystyle{splncs}
\bibliography{library}
\end{document}